%% file: main.tex
\pdfoutput=1

\documentclass[11pt]{article}

\usepackage[final]{acl}

\usepackage{times}
\usepackage{booktabs}
\usepackage{latexsym}
\usepackage{amsmath}
\usepackage{bm}
\usepackage{dsfont}

\usepackage{amssymb}
\usepackage{pifont}

\usepackage{makecell}
\usepackage{multirow}
\usepackage[T1]{fontenc}

\usepackage[utf8]{inputenc}

\usepackage{microtype}
\usepackage{graphicx}
\usepackage[most]{tcolorbox}
\usepackage{inconsolata}

\newcommand{\abr}[1]{\textsc{#1}}
\newcommand{\snli}{\abr{Snli}}
\newcommand{\dnli}{$\delta$-\abr{nli}}
\newcommand{\dsnli}{$\delta$-\snli}
\newcommand{\dsnlitest}{\dsnli-\abr{test}}
\newcommand{\snlitest}{\snli-\abr{test-sample}}
\definecolor{lightteal}{RGB}{234,209,220}
\definecolor{teal}{RGB}{116,27,71}
\definecolor{lightblue}{RGB}{181, 179, 242}
\definecolor{darkblue}{RGB}{68, 63, 204}
\newcommand{\cmark}{\ding{51}}%
\newcommand{\xmark}{\ding{55}}%

\newenvironment{tight_enumerate}{
\begin{enumerate}
  \setlength{\itemsep}{0pt}
  \setlength{\parskip}{0pt}
}{\end{enumerate}}

\newtcolorbox[list inside=prompt,auto counter,number within=section]{prompt}[1][]{
    colbacktitle=black!60,
    fonttitle=\small,
    coltitle=white,
    fontupper=\footnotesize,
    boxsep=4pt,
    left=0pt,
    right=0pt,
    top=0pt,
    bottom=0pt,
    boxrule=1pt,
    #1,
}

\newcommand{\bucket}{\includegraphics[height=10pt,width=10pt]{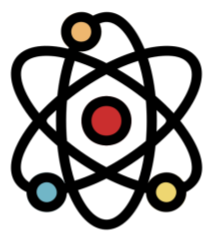}}
\newcommand{\bucketsmall}{\includegraphics[height=9pt,width=9pt]{figures/atom.png}}

\newenvironment{packed_itemize}{
\begin{itemize}
  \setlength{\itemsep}{1pt}
  \setlength{\parskip}{0pt}
  \setlength{\parsep}{0pt}
}{\end{itemize}}

\definecolor{lightgrey}{RGB}{235, 236, 237}
\definecolor{darkgrey}{RGB}{124, 124, 125}

\definecolor{lightteal}{RGB}{208,223,226}
\definecolor{teal}{RGB}{69,129,129}

\definecolor{lightorange}{RGB}{252,229,205}
\definecolor{burntorange}{RGB}{207,146,82}

\definecolor{lightpurple}{RGB}{217,210,233}
\definecolor{darkpurple}{RGB}{124,102,179}
\definecolor{coolgreen}{RGB}{73, 176, 104}
\definecolor{coolred}{RGB}{235, 125, 120}

\title{NLI under the Microscope:\\ What Atomic Hypothesis Decomposition Reveals}

\author{Neha Srikanth \\
  University of Maryland, College Park \\
  \texttt{nehasrik@umd.edu} \\\And
  Rachel Rudinger \\
  University of Maryland, College Park \\
  \texttt{rudinger@umd.edu} \\}

\begin{document}
\maketitle
\begin{abstract}
\input{sections/00-abstract}
\end{abstract}

\section{Introduction}

\input{sections/10-intro}

\section{Background}
\label{sec:task-background}
\input{sections/20-background}

\section{Atomic Sub-Problems in NLI}
\label{sec:atom-generation}

\input{sections/30-atoms}

\section{Atoms in Traditional NLI}
\label{sec:snli-atoms}
\input{sections/40-snli}

\section{Atoms in Defeasible Inference}
\label{sec:dnli-atoms}
\input{sections/50-dnli}

\section{Measuring Inferential Consistency in \dnli}
\label{sec:diversity}
\input{sections/55-diversity}

\section{Conclusion}
\input{sections/70-conclusion}

\section*{Limitations}
This paper uses LLMs to generate atomic decompositions of hypotheses. 
While we did validate whether or not generated atoms were \textit{valid} during manual annotation for \dsnli, we did not determine whether each family had missing decompositions of a particular granularity for \snlitest. 

Both \dnli~and \snli~have been shown to contain annotation artifacts~\cite{gururangan-etal-2018-annotation}, or particular statistical patterns between hypotheses (in the case of \snli) and updates (in the case of \dnli). 
While prompt-based models are not directly trained on any of the datasets, the performance of finetuned \texttt{roberta-large} and \texttt{deberta-v3-large} may be inflated by the presence of annotation artifacts in update sentences. 
However, our methodology provides an opportunity for the collection of updates free from these artifacts by collecting updates that indirectly or lightly affect on hypotheses.

Finally, since the original train set of \dnli~does not contain ``no effect'' or ``neutral'' updates, we report metrics for \textit{finetuned} models on the subset of non-neutral atomic sub-problems.
Future work may augment \dnli~finetuning data with \textit{contextual} neutral atomic sub-problems (akin to hard negatives in information retrieval) to enable this predictive ability in finetuned models.

\section*{Acknowledgments}
We thank the anonymous reviewers as well as the members of the University of Maryland CLIP lab for their thoughtful and thorough feedback.
We also thank Joe Stacey for feedback on an earlier draft of this paper.
This work was supported by NIH Award No. R01MD016037 (Srikanth) and NSF CAREER Award No. 2339746 (Rudinger).
The content is solely the responsibility of the authors and does not necessarily represent the official views of the National Institutes of Health or the National Science Foundation. 
The funders had no role in study design, data collection and analysis, decision to publish, or preparation of the paper.

\bibliography{custom}

\clearpage
\appendix
\input{sections/appendix}

\end{document}

%% file: sections/00-abstract.tex
Decomposition of text into atomic propositions is a flexible framework allowing for the closer inspection of input and output text.
We use atomic decomposition of hypotheses in two natural language reasoning tasks, traditional NLI and defeasible NLI, to form \textit{atomic sub-problems}, or granular inferences that models must weigh when solving the overall problem.
These atomic sub-problems serve as a tool to further understand the structure of both NLI and defeasible reasoning, probe a model's consistency and understanding of different inferences, and measure the diversity of examples in benchmark datasets. 
Our results indicate that LLMs still struggle with logical consistency on atomic NLI and defeasible NLI sub-problems.
Lastly, we identify \textit{critical atomic sub-problems} of defeasible NLI examples, or those that most contribute to the overall label, and propose a method to measure the \textit{inferential consistency} of a model, a metric designed to capture the degree to which a model makes consistently correct or incorrect predictions about the same fact under different contexts.

%% file: sections/10-intro.tex
Atomic decomposition involves breaking sentences down into \textit{atomic propositions}, or granular facts that are explicitly supported by the original text.
This style of decomposition has widespread applications, including assessing the factual precision of generated text~\cite{min-etal-2023-factscore}, claim verification~\cite{chen-etal-2024-complex}, and multihop QA~\cite{perez-etal-2020-unsupervised}, since it allows for careful, finer-grained inspection of text.

We use atomic decomposition as tool to dive deeper into two types of natural language reasoning: traditional NLI~\cite{giampiccolo2007third} and defeasible inference~\cite{rudinger-etal-2020-thinking}, a mode of reasoning where inferences may change in light of new evidence.
In both tasks, atomic decomposition of hypotheses into \textit{atoms} breaks complex sentences into granular pieces of information that models must weigh when drawing higher level inferences, producing \textit{atomic sub-problems}.
We use these sub-problems not only for better insight into the structure and nuances of NLI and defeasible NLI, but also to assess accompanying benchmarks and to more deeply probe the robustness of models' situational understanding.\footnote{Code and data available at \url{https://github.com/nehasrikn/nli-atoms}.}

\begin{figure}[t!]
\centering
\includegraphics[scale=0.715]{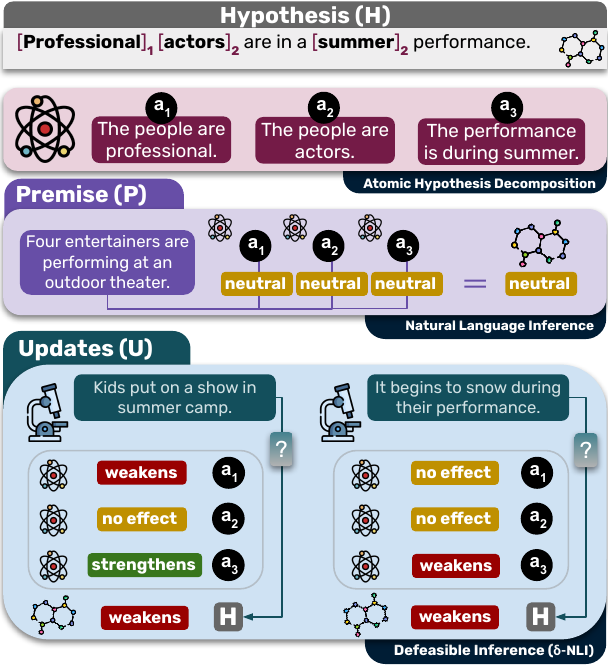}
\caption{Top: Atomic hypothesis decomposition breaks down hypotheses ($H$) into  entailed propositional ``atoms'' ($a_1-a_3$). Middle: Pairing the premise ($P$) with each atom yields a set of NLI sub-problems ($P+a$); the sub-problem labels predict the full NLI problem ($P+H$) label. Bottom: Paired with an update ($U$), each atom yields a defeasible NLI sub-problem ($P+a+U$); the set of sub-problem labels are predictive of the full problem ($P+H+U$) label, but the non-monotonic relationship is more complex than for traditional NLI.}
\label{fig:teaser-example}
\end{figure}

Consider the example in Figure~\ref{fig:teaser-example} of a premise and hypothesis with a neutral relationship. 
When predicting an overall neutral relation, a model must determine the relation between the premise and each of the three atomic propositions that together form the hypothesis --- in this case, all of which are neutral.
If the model predicts an entailment or a contradiction between the premise and one of the atoms, its understanding of the situation may be called into question.

We first present our two tasks of interest (\S\ref{sec:task-background}) and discuss the utility of atomic sub-problems and their construction (\S\ref{sec:atom-generation}).
Then, we analyze the behavior of large language models (LLMs) on atomic sub-problems in traditional NLI (\S\ref{sec:snli-atoms}) where we evaluate their logical consistency between each original \snli~instance and its corresponding sub-problems.
We find that despite high accuracy, LLMs still struggle with logical consistency.
Then, we study atomic sub-problems in defeasible NLI (\S\ref{sec:dnli-atoms}) and propose a framework to pinpoint the inference(s) evaluated in each example by way of the \textit{question under discussion} of examples, a well-studied linguistic phenomenon~\cite{benz2017questions, wu-etal-2023-qudeval}.
Finally, we present a method to group defeasible NLI examples based on related atomic sub-problems (\S\ref{sec:diversity}) and measure the \textit{inferential consistency} of a model, a metric capturing the likelihood that its prediction for a particular inference will remain \textbf{consistently} correct or incorrect under different contexts.

%% file: sections/20-background.tex
\subsection{Natural Language Inference}
Natural language inference~\cite{giampiccolo2007third, maccartney2009natural, bowman-etal-2015-large} is the task of determining whether a premise $P$ entails, contradicts, or is neutral in relation to a hypothesis, $H$.
For example, the premise \textit{``A little girl in a lush green field''} contradicts the hypothesis \textit{``A little girl rides her ox in a desert''}, as ``desert'' directly contradicts the ``lush green field''.

\paragraph{\snli.}
The first large-scale NLI dataset, \snli, uses image captions as premises paired with hypotheses elicited from crowdworkers.
Though this benchmark has been well-saturated by LLMs over the past few years, it serves as a useful resource for studies in model robustness~\cite{srikanth-rudinger-2022-partial, kaushik2019learning} and annotation artifacts~\cite{gururangan-etal-2018-annotation}.

\subsection{Defeasible Inference}

Defeasible reasoning is a form of non-monotonic reasoning in which inferences may be altered in light of new evidence~\cite{reiter1980logic}.
For example, given the premise \textit{``A group of people sitting around a rectangular table''}, the inference \textit{``they have a work meeting''} is weakened upon learning that \textit{``the people are toddlers.''}

\paragraph{\dnli.}
\citet{rudinger-etal-2020-thinking} introduce the task of defeasible natural language inference and an accompanying benchmark, \dnli. 
Given a $P$ and $H$ pair with a \textit{neutral} relation, a third \textbf{update} sentence $U$ \textit{strengthens} $H$ if, upon reading $U$, $H$ is more likely to be true, and \textit{weakens} $H$ if $H$ is less likely to be true upon reading $U$.
Defeasible NLI is then a binary classification task of predicting a strengthener or weakener label for a $(P, H, U)$ set.
\dnli~is built on top of three popular commonsense reasoning datasets: \snli~\cite{bowman-etal-2015-large}, \abr{atomic}~\cite{sap2019atomic}, and \abr{social-chem-101}~\cite{forbes2020social}.
For a $P$ and $H$ pair (or, in the case of \abr{social-chem-101}, just $H$), crowdworkers write an update sentence for a target label, but are not instructed to target a particular part of $H$ when doing so.
The authors ensure that the train, development, and test splits of the data are split at the $P$-$H$ level to avoid leakage.
Crowdworkers may not write updates that directly contradict information in the premise.
For simplicity, we focus on the \snli-derived split of \dnli, or \dsnli, which selects neutral $P$-$H$ \snli~pairs.

\subsection{Related Work}
Atomic decomposition has been used in fact checking~\cite{min-etal-2023-factscore, glover-etal-2022-revisiting, yuan-vlachos-2024-zero}, claim verification~\cite{chen-etal-2024-complex}, summarization~\cite{nenkova-passonneau-2004-evaluating}, and text-to-image generation~\cite{cho2023davidsonian} among others.
\citet{kamoi-etal-2023-wice} construct a dataset of claims and sub-claims for claim checking where sub-claims, analogous to our atoms, are labeled with respect to evidence.
Most relevant to our work, \citet{stacey-etal-2022-logical} train a span-based NLI model to make \textit{span-level} decisions on \snli~and SICK examples that are composed to produce an overall label. 
In contrast, our work measures the logical consistency of LLMs that may have seen NLI data during pretraining, but that are not explicitly trained to weigh atomic inferences.
Their followup~\cite{stacey-etal-2024-atomic} \textit{trains} another NLI system using LLM-generated atoms, however their study focuses primarily on \textit{premise} decomposition.
To the best of our knowledge, our study is the first to explore atoms in defeasible inference.

%% file: sections/30-atoms.tex
\input{tables/examples}

Reasoning about situations often involves weighing multiple pieces of information to draw inferences. 
Consider the last row in Table~\ref{table:atom-examples}. 
A human determining that $H$ contradicts $P$ will attribute the contradiction to the fact that $P$ mentions a father and daughter, but $H$ mentions two men.
Implicitly, they will have also weighed the fact that $H$'s mention of ``cutting grass'' is entailed by $P$'s mention of a lawnmower, and so it does not contribute to the contradiction.
We treat these two determinations as distinct \textit{atomic sub-problems}.

Hypotheses in \snli, and in turn, \dsnli, can be complex sentences, and while solving inference problems, models must weigh all pieces of information in both $P$ and $H$.
We expect humans to make inferences about constituent pieces of information in a manner that is consistent with their overall judgment, an equally desirable property in models.
Not only does it signal holistic understanding of the situation described in the problem, but it can help pinpoint exactly what types of inferences models struggle with.
Identifying atomic sub-problems also allows us to understand the granular inferences that are evaluated in benchmark datasets.
In turn, this helps to understand the \textit{diversity} in the dataset: despite there being thousands of examples, certain inferences may come up repeatedly.

To identify the constituent sub-problems, we break \textit{hypotheses} in \snli~and \dsnli~into atomic propositions~\cite{wanner2024closer} to use in subsequent analyses. 
Each atomic decomposition represents a single piece of information.
Formally, given an \snli~example with $P$ and $H$, we generate atomic decompositions of $H$ represented by $a_1...a_n$. 
Each atomic sub-problem then involves predicting the relation between a ($P$, $a_i$) tuple (\S\ref{sec:snli-atoms:rules}).
Given a \dnli~example with $P$, $H$, and $U$, atomic sub-problems involve a ($P$, $a_i$, $U$) tuple where the task is to determine whether $U$ strengthens, weakens, or has no effect on $a_i$ (\S\ref{sec:dnli-atoms}).

\subsection{Generating Atomic Propositions}
\label{subsec:generate-atoms}
To generate atomic propositions, we draw on Neo-Davidsonian event-based semantic representations of sentences~\cite{castaneda1967, parsons1990events}.
Sentences can be represented in first-order logical form as conjunctions of predicates representing entities, where actions are explicitly represented with event variables and predicate arguments are mapped to semantic roles~\cite{dowty1991thematic}. 
For example, the sentence \textit{``The juggler performs at a party''} could be represented as: 
\[
\small
\begin{aligned}
    &\exists x_1 \exists e \ (\text{Juggler}(x_1) \land \text{Perform}(e) \land \text{Agent}(e, x_1) \land \\
    &\quad \exists x_2 \ (\text{Party}(x_2) \land \text{At}(e, x_2)))
\end{aligned}
\]

\noindent Each conjunct can then be mapped to a natural language expression, called an \textit{atom}.
This ensures that both arguments of actions \textit{and} the actions themselves are included as separate atoms.

We draw on this intuition to carefully hand-construct exemplars, a methodology shown to improve the atomicity and groundedness of decompositions~\cite{wanner2024closer}.
We prompt \texttt{llama-3-8b-instruct} with these exemplars (Appendix~\ref{appendix:atom-generation}) to generate atoms for each example in the \dsnli~test set (henceforth, \dsnlitest),  as well as for 1000 randomly sampled examples in the SNLI test set (\snlitest). See Table~\ref{tab:dataset-sizes} for dataset statistics.

\subsection{Validating Atomic Decompositions}
\label{subsec:validate-atoms}
Valid atomic decompositions of hypotheses must be logically entailed from the hypothesis they were decomposed from.
For our experiments on \snlitest~(\S\ref{sec:snli-atoms}), we do not validate atom entailment ourselves, letting each model determine whether $H$ entails each $a_i$ itself (\S\ref{sec:snli-atoms:rules}), and only measuring consistency on the atomic sub-problems that the model itself admits as ``valid''.

However, we \textit{do} validate all generated atoms in \dsnlitest, since non-monotonic reasoning does not give rise to clear constraints between an original problem and its constituent atomic sub-problems.
Our two-step validation process involves pruning decompositions with a strong, finetuned NLI model followed by human validation.

\paragraph{Pruning.} For each example in \dsnlitest, we use a \abr{DeBERTa}-large model finetuned on popular NLI datasets\footnote{MNLI~\cite{williams-etal-2018-broad}, Fever-NLI~\cite{thorne-etal-2018-fever}, Adversarial NLI~\cite{nie-etal-2020-adversarial}, LingNLI~\cite{parrish-etal-2021-putting-linguist}, and WANLI~\cite{liu-etal-2022-wanli}} and remove all generated atoms that are not entailed by the hypothesis.
By design, $P$-$H$ pairs in \dsnli~have a neutral relation, and updates strengthen or weaken propositions in the \textit{hypothesis}.
Hence, we run a secondary pruning stage to retain only those atoms that are \textit{not} entailed by the premise (see Table~\ref{tab:dataset-sizes}).
See Appendix~\ref{appendix:atom-generation} for a discussion of coverage.
\paragraph{Human Validation.} An author annotated all atoms that survived pruning as either \textit{invalid} or \textit{valid} (see Table~\ref{table:atom-examples} for examples of invalid atoms).
Valid \dsnli~atoms had to (1) be grammatical, (2) entail from $H$, (3) not entail from $P$.
Atoms introducing new information were considered invalid, including those that were \textit{pragmatic} inferences of $H$~\cite{jeretic-etal-2020-natural, srikanth-etal-2024-pregnant}.

95.7\% of pruned atoms were determined as valid by the author annotator.
An external annotator also annotated a sample of 100 atoms for validity for an agreement of $\kappa=0.82$ measured by Cohen's Kappa~\cite{cohen1960coefficient}.
The remaining analysis in this work is done on the set of \textit{valid} \dsnli~atoms.

%% file: tables/examples.tex
\begin{table*}[ht!]
\centering
\resizebox{2.08\columnwidth}{!}{
    \begin{tabular}{p{0.25\linewidth}p{0.18\linewidth}p{0.2\linewidth}p{0.65\linewidth}} 
    \toprule
    \multicolumn{1}{c}{\textbf{\textbf{Premise ($P$)}}} & \multicolumn{1}{c}{\textbf{Hypothesis ($H$)}} & \multicolumn{1}{c}{\textbf{Update ($U$)}} & \multicolumn{1}{c}{\textbf{Atoms}} \\ 
    \midrule
    A man in a white t-shirt and jeans is holding a mallet and chisel next to his abstract sculpture which stands on several bricks. &
    A man is trying to finish his sculpture for a church &
    The man has taken his first strike against the granite.\newline(Weakener) &
    \colorbox{lightteal}{\textcolor{teal}{\textbf{$a_1$}:}} \textbf{The thing the person is trying to do is finish. (-1) }\newline
    \colorbox{lightteal}{\textcolor{teal}{\textbf{$a_2$}:}} The thing the person is trying to finish is a sculpture. (+1) \newline
    \colorbox{lightteal}{\textcolor{teal}{\textbf{$a_3$}:}} The sculpture is for a church. (0)
    \\
    \midrule
    There is a green trash truck in road with a person sweeping sidewalk. &
    The garbage man sweeps up where the can spilled. &
    The person is wearing a city uniform.\newline(Strengthener) &
    \colorbox{lightteal}{\textcolor{teal}{\textbf{$a_1$}:}} \textbf{The person is a garbage man. (+2)} \newline
    \colorbox{lightteal}{\textcolor{teal}{\textbf{$a_2$}:}} The thing being swept is up. (invalid) \newline
    \colorbox{lightteal}{\textcolor{teal}{\textbf{$a_3$}:}} There is a can. (0) \newline
    \colorbox{lightteal}{\textcolor{teal}{\textbf{$a_4$}:}} The can has spilled. (+1) \newline
    \colorbox{lightteal}{\textcolor{teal}{\textbf{$a_5$}:}} \textbf{The person is sweeping up a spill. (+2)}
    \\
    \toprule
    Two young men climb a tree overlooking a rural setting, with one of them out far on a limb and clutching a white helmet. &
    Two brothers are climbing a tree to get down their Frisbee. (N) &
    N/A &
    \colorbox{lightteal}{\textcolor{teal}{\textbf{$a_1$}:}} There are two people. (E) \newline
    \colorbox{lightteal}{\textcolor{teal}{\textbf{$a_2$}:}} There are two people who are brothers. (N) \newline
    \colorbox{lightteal}{\textcolor{teal}{\textbf{$a_3$}:}} There are people climbing. (E)\newline
    \colorbox{lightteal}{\textcolor{teal}{\textbf{$a_4$}:}} There are people climbing a tree. (E) \newline
    \colorbox{lightteal}{\textcolor{teal}{\textbf{$a_5$}:}} There is a purpose for people climbing a tree. (N) \newline
    \colorbox{lightteal}{\textcolor{teal}{\textbf{$a_6$}:}} The purpose for people climbing a tree is to get something. (N)\newline
    \colorbox{lightteal}{\textcolor{teal}{\textbf{$a_7$}:}} The thing people are trying to get is a Frisbee. (N)\newline
    \colorbox{lightteal}{\textcolor{teal}{\textbf{$a_6$}:}} The thing people are trying to get is down. (invalid)
    \\
    \toprule
    A father and his daughter are riding a lawn mower down a street while dressed in American colors. &
    two men cut grass by hand (C) &
    N/A &
    \colorbox{lightteal}{\textcolor{teal}{\textbf{$a_1$}:}} There are two people. (E) \newline
    \colorbox{lightteal}{\textcolor{teal}{\textbf{$a_2$}:}} There are two people who are men. (C) \newline
    \colorbox{lightteal}{\textcolor{teal}{\textbf{$a_3$}:}} There are people cutting. (E)\newline
    \colorbox{lightteal}{\textcolor{teal}{\textbf{$a_4$}:}} There are people cutting grass. (E) \newline
    \colorbox{lightteal}{\textcolor{teal}{\textbf{$a_5$}:}} There is a method of cutting grass. (E) \newline
    \colorbox{lightteal}{\textcolor{teal}{\textbf{$a_6$}:}} The method of cutting grass is by hand (C)\\
    \bottomrule
    \end{tabular}
}
\caption{\dsnli~(Rows 1 and 2) and SNLI (Rows 3 and 4) instances, along with atomic decompositions of hypotheses and the label of the sub-problem involving that atom.}
\label{table:atom-examples}
\end{table*}

%% file: sections/40-snli.tex
\input{tables/benchmarking_snli}

While performing strongly on various benchmarks, LLMs still struggle with many types of consistency including paraphrastic consistency~\cite{srikanth2024often, verma-etal-2023-evaluating}, hypothetical consistency~\cite{chen2023two}, or even preferential consistency~\cite{zhao2024measuring}.
When LLMs make entailment judgments, another desirable property is \textit{logical consistency}.
Namely, when an LLM itself deems a set of atoms $a_1$...$a_n$ entailed by $H$, we can hold it accountable to maintain consistency between its judgments on each ($P$, $a_i$) sub-problem and its overall ($P$, $H$) judgment in a logical way.
This gives us necessary, but not sufficient, evidence to help signal that it has ``understood'' the situation.

\subsection{Atomic and Overall Label Consistency}
\label{sec:snli-atoms:rules}
We construct a set of rules to establish the relationship between atomic sub-problems and overall problem labels.
\vspace{-0.5em}
\begin{tight_enumerate}
    \item If $H$ is entailed by $P$: Each valid $a_i$ must be entailed by $P$.
    \item If $H$ contradicts $P$: At least one valid $a_i$ must contradict $P$.
    \item If $H$ is neutral with respect to $P$: At least one valid $a_i$ must be neutral with respect to $P$, all others may be either neutral or entailed.
\end{tight_enumerate}

\subsection{Experimental Setup}
We experiment with six LLMs: \texttt{gpt-4o}~\cite{openai2024gpt4o}, \texttt{gpt-4o-mini}~\cite{openai2024gpt4o}, \texttt{gpt-3.5-turbo-0125}~\cite{ouyang2022training}, \texttt{llama-3-8b-instruct}~\cite{dubey2024llama}, \texttt{llama-3-70b-instruct}~\cite{dubey2024llama}, and \texttt{gemma-2-9b-instruct}~\cite{team2024gemma}.

First, we benchmark each models's performance on \snlitest~original examples (Table~\ref{tab:benchmarking-snli}) using Prompt~\ref{prompt:nli} adapted from~\citet{liu2023evaluating}.
We use 12 in-context \textit{original} \snli~examples from the dev split evenly distributed over the three NLI labels.
Then, for each \snlitest~example, we have each model predict the relation between $H$ and each generated atom $a_i$ from~\S\ref{subsec:generate-atoms} using the same prompt and exemplar set.
For each atom that the LLM predicts as entailed, we have it predict the relation between $P$ and $a_i$ using the same prompt and exemplar set.

We report overall logical consistency as the percent of examples where the full prediction was logically consistent with the predicted labels for sub-problems as dictated by the rules in \S\ref{sec:snli-atoms:rules}.

\paragraph{Results.} Despite higher \textit{accuracy} numbers, LLMs seem to struggle with \textit{logical consistency} (Table~\ref{tab:benchmarking-snli}).
Interestingly, a model's accuracy is not fully indicative of its logical consistency. 
When models incorrectly predict the full example's label, they are more prone to logical inconsistencies between atomic sub-problems and the full problem (Table~\ref{tab:benchmarking-snli}, Columns 3 and 4).
Though not the top performing model, \texttt{gpt-4o} outperformed other models on logical consistency \textbf{even on examples where its full prediction was incorrect}.
Such logical consistency indirectly captures the \textit{reliability} of a model's full prediction: when two LLMs achieve similar accuracies, logical consistency serves as another point of comparison.

We also stratify our results by the predicted overall label and report logical consistency within each class (Table~\ref{tab:benchmarking-snli}, Columns 5---7). 
All models exhibit consistency gaps \textit{within the 3 labels}, and different models struggle with different example classes.

Lastly, we experiment with \textit{atomic inference}~\cite{stacey-etal-2024-atomic}, or inducing an overall label via logical rules over the predicted atomic sub-problem labels, to understand whether a setting in which models only provide granular inferences can be more effective.
We induce an overall label with similar logical rules to those in \S\ref{sec:snli-atoms:rules}: (1) if all $a_i$ are predicted as entailed by $P$, we predict entailment, (2) if at least one $a_i$ is predicted as contradicting $P$, we predict a contradiction, (3) otherwise, predict neutral.
While this strategy does not yield competitive performance with full example accuracy (Table~\ref{tab:benchmarking-snli}, Column 1 versus Column 8), it does offer a more \textit{interpretable} framework for LLMs that otherwise seem to struggle with logical consistency.
Atom judgments may more difficult than overall judgments for a variety of reasons (see Appendix~\ref{appendix:inconsistency-analysis}), and in turn, inducing a label from individual atomic predictions may be less reliable.

%% file: tables/benchmarking_snli.tex
\begin{table*}
\centering
\resizebox{2.08\columnwidth}{!}{
\setlength\tabcolsep{2pt}
\begin{tabular}{lcccccccc} 
\toprule
\multirow{2}{*}{}      & \multirow{2}{*}{\begin{tabular}[c]{@{}c@{}}\textbf{Full Example}\\\textbf{Accuracy}\end{tabular}} & \multirow{2}{*}{\begin{tabular}[c]{@{}c@{}}\textbf{Overall Logical }\\\textbf{Consistency}\end{tabular}} & \multirow{2}{*}{\begin{tabular}[c]{@{}c@{}}\textbf{Consistency on}\\\textbf{Correct Exs}\end{tabular}} & \multirow{2}{*}{\begin{tabular}[c]{@{}c@{}}\textbf{\textbf{Consistency on}}\\\textbf{\textbf{Incorrect Exs}}\end{tabular}} & \multicolumn{3}{c}{\textbf{Logical Consistency by Label}}       & \multirow{2}{*}{\begin{tabular}[c]{@{}c@{}}\textbf{Induced Atom}\\\textbf{Label Accuracy}\end{tabular}}  \\
                       &                                                                                                   &                                                                                                          &                                                                                                        &                                                                                                                            & \texttt{entailment} & \texttt{neutral} & \texttt{contradiction} &                                                                                                          \\ 
\midrule
\texttt{gpt-4o-mini-2024-07-18} & $\mathbf{89.8}$                                                                                            & $84.0$                                                                                                   & $86.8$                                                                                                 & $59.8$                                                                                                                     & $78.8$              & $90.1$           & $82.6$                 & $81.3$                                                                                                   \\
\texttt{gpt-4o-2024-08-06}      & $88.5$                                                                                            & $\mathbf{87.9}$                                                                                                   & $\mathbf{89.6}$                                                                                                 & $\mathbf{74.8}$                                                                                                                     & $88.9$              & $92.7$           & $79.2$                 & $81.9$                                                                                                   \\
\texttt{llama-3-70b-it}         & $87.7$                                                                                            & $84.2$                                                                                                   & $89.5$                                                                                                 & $46.3$                                                                                                                     & $82.1$              & $81.3$           & $89.5$                 & $\mathbf{84.8}$                                                                                                   \\
\texttt{llama-3-8b-it}          & $85.2$                                                                                            & $81.2$                                                                                                   & $88.2$                                                                                                 & $41.2$                                                                                                                     & $76.0$              & $85.0$           & $85.1$                 & $82.9$                                                                                                   \\
\texttt{gemma-2-9b-it}          & $84.2$                                                                                            & $80.9$                                                                                                   & $85.9$                                                                                                 & $54.4$                                                                                                                     & $84.0$              & $82.4$           & $73.8$                 & $78.9$                                                                                                   \\
\texttt{gpt-35-turbo-0125}      & $82.1$                                                                                            & $74.4$                                                                                                   & $80.4$                                                                                                 & $46.9$                                                                                                                     & $70.4$              & $90.9$           & $64.2$                 & $74.9$                                                                                                   \\
\bottomrule
\end{tabular}}
\caption{Accuracy (col. 1) and logical consistency of LLMs full SNLI examples (col. 2). We also report logical consistency on examples where the full prediction was correct (col. 3) and incorrect (col 4.), as well as stratify by predicted label (cols. 5--7). Finally, we logically compose atomic sub-problem labels to see if this more interpretable method outperforms full example accuracy (col. 8).}
\label{tab:benchmarking-snli}
\end{table*}

%% file: sections/50-dnli.tex
We now turn to defeasible inference to explore how atomic sub-problems can help us better understand the complexities of the task, model performance, and the knowledge evaluated in the \dsnli~dataset.

\begin{figure*}[t!]
\centering
\includegraphics[width=\textwidth]{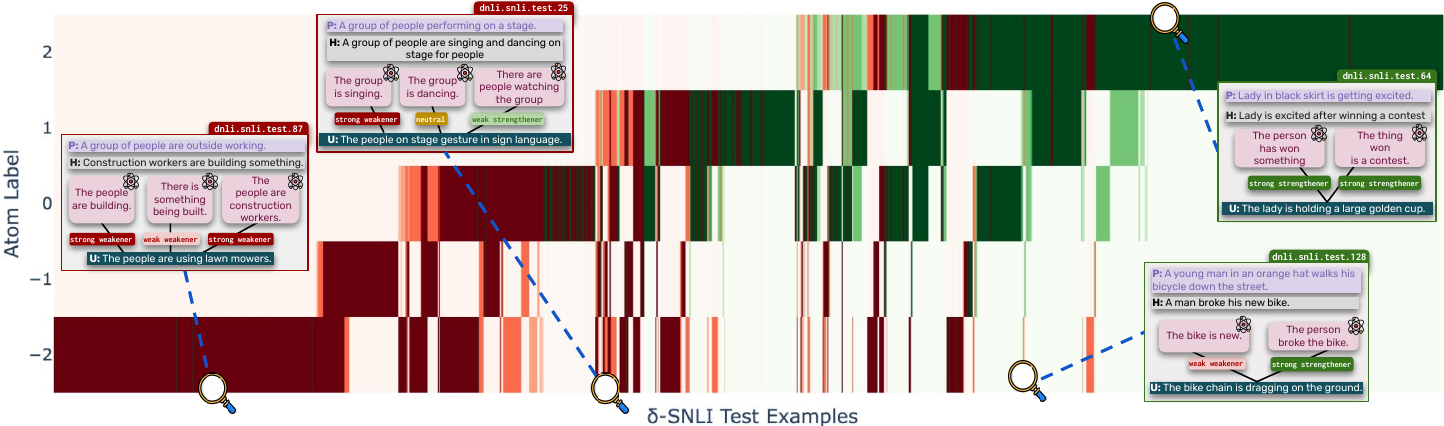}
\caption{A rug plot visualization of 1,761 \dsnli~instances and their corresponding distribution of atomic sub-problem labels. Each vertical slice represents one full \dsnli~instance. Slice color (red or green) represents the full instance label (weakener or strengthener). For each \dsnli~problem, we manually label each corresponding atomic sub-problem on a -2 (strongly weakens) to +2 (strongly strengthens) scale. Each vertical slice uses shading (light/dark) to represent the resulting distribution of atomic sub-problem labels (-2 to +2). Slices are ordered left to right by proportion of weakener labels, showing relatively high separation between red and green instances. When atomic sub-problems contain a mix of positive and negative labels, the full problem label may be a strengthener or a weakener, as illustrated by the two center-most exemplars.}

\label{fig:label-distribution}
\end{figure*}

In traditional NLI (\S\ref{sec:snli-atoms}), labels of atomic sub-problems function like terms in an equation---the overall relation between $P$ and $H$ can be computed from strict logical rules over relations between $P$ and individual $a_i$.
In contrast, defeasible inference functions akin to fuzzy logic~\cite{castro1998non}.
Determining the overall effect of the update $U$ on $H$ involves a softer weighing of the direction and magnitude of its effect on each atom.

For example, consider the second row in Table~\ref{table:atom-examples}. 
A human reading the update $U$ (\textit{``The person is wearing a city uniform''}) would conclude that $U$ \textit{strengthens} their belief in $H$ (\textit{``The garbage man sweeps up where the can spilled''}) as opposed to weakens it.
Looking at this problem through the lens of atomic decomposition helps pinpoint why.
$H$ consists of five atoms, each representing pieces of evidence \textit{not} present in $P$ ripe for targeting by updates.
Two of the five atoms are most strongly supported by $U$: the person is wearing a city uniform strongly strengthens our belief that they may be a garbage man ($a_1$) as well as that they may be sweeping up a spill ($a_5$). 
However, $U$ has no effect on our belief that there is a can in the scenario ($a_3$).
The co-occurrence of $a_1$ and $a_5$ compound the strengthening effect, leading to an overall strengthening effect of $U$ on $H$.

Breaking down \dsnli~hypotheses and forming atomic sub-problems in this manner gives us a framework to understand the intricacies of defeasible inference.
We begin by benchmarking recent LLMs to understand the state of defeasible inference capabilities of models.
Then, we introduce the idea of a \textit{critical atom}, or the primary piece of information an update acts on.
Finally, we use critical atoms as a way to better understand and interpret model behavior, as well as argue that critical atoms serve as a useful representation for measuring the type of knowledge evaluated in \dsnli.

\subsection{Understanding Defeasible Inference with Atomic Sub-Problems}

\paragraph{Benchmarking LLMs on \dnli.} We benchmark a suite of recent models on full examples from \dsnlitest, including encoder models and prompt-based models, open-source and proprietary systems, as well as models of various sizes. 
We finetune encoder-only models (\texttt{roberta-large} and \texttt{deberta-v3-large}) on the train set of \dsnli~for 2 epochs with a learning rate of 2e-5 and a batch size of 32.
For all prompt-based models, we do few-shot evaluation with Prompt~\ref{prompt:defeasible} and 10 in-context examples evenly split between strengtheners and weakeners.

Many of the models in our suite surpass the human performance benchmarked by~\citet{rudinger-etal-2020-thinking}, with \texttt{gpt-4o} as the top performing model at 92\% accuracy (Table~\ref{tab:benchmarking-dnli}).
However, since it remains unclear whether this accuracy is indicative of a holistic understanding of situations in \dsnli, we turn to studying performance on the atomic reasoning problems that compose each \dsnli~example to better contextualize these results.
\paragraph{Annotating Atoms.} Each atomic sub-problem in \dsnli~is $(P, a_i, U)$ tuple capturing the effect of the update on a specific atom $a_i$. 
An author annotated all valid (as determined in \S\ref{subsec:validate-atoms}) atomic sub-problems for each example in \dsnlitest~according to the five-point scale used in~\citet{rudinger-etal-2020-thinking} for validation (Table~\ref{tab:dataset-sizes}) ranging from \textit{strongly weakens} (-2) to \textit{strongly strengthens} (+2)  with a midpoint value of \textit{no effect} (0) for atoms on which $U$ had no effect. 
The same external annotator from \S\ref{subsec:validate-atoms} annotated a random sample of 100 \textit{valid} atomic sub-problems on the same -2 to +2 scale (Appendix~\ref{appendix:instructions}), obtaining an agreement of $\tau=0.79$ with Kendall's Tau~\cite{kendall1938new}.
\input{tables/benchmarking_dnli}
\paragraph{Ground Truth Label Distribution.} Figure~\ref{fig:label-distribution} visualizes the label distribution ($-2$ to $+2$) of atomic sub-problems of each \dsnlitest~example as a thin vertical strip. Strips are green if the original example is a strengthener and red for weakeners.
While some examples have all atomic labels of the \textit{same} polarity, \textbf{a significant chunk of the dataset includes atomic sub-problems with no effect or the \textit{opposite} polarity}.
Examples depicted across the spectrum in Figure~\ref{fig:label-distribution} illustrate the non-monotonicity of defeasible inference. 

\paragraph{Atomic Sub-Problem Performance.} 
We first measure the performance of models on atomic sub-problems using annotated ground-truth labels.
The original \dsnli~dataset was designed as a binary prediction task. 
However, as Figure~\ref{fig:label-distribution} depicts, updates may also have no effect on atoms.
We adapt Prompt~\ref{prompt:defeasible} to accommodate this ternary task (Prompt~\ref{prompt:defeasible-atom}) and use atoms in exemplars instead of full hypotheses.
%
Since the train set of \dnli~only admits binary labels, we reuse the finetuned models (\texttt{deberta} and \texttt{roberta}), and report atom accuracy on non-neutral atoms (80\% of atoms, Figure~\ref{fig:atom-label-hist}). 

Across the board, models perform worse on atomic sub-problems than on full examples (Table \ref{tab:benchmarking-dnli}, Column 2).
Since updates often act on multiple parts $U$ (Table~\ref{table:atom-examples}), we hypothesize that this compounding effect may contribute to higher performances on full examples.

We observe that some atomic sub-problems are more critical contributors to the overall effect of $U$ on $H$.
Consider the example in row 2 of Table~\ref{table:atom-examples}.
Since $U$ acts most strongly on $a_1$ and $a_5$, \textbf{we can assume that they are critical in determining the overall effect of $U$ on $H$}, and correctly understanding the effect of $U$ on $a_3$ or $a_4$ is not essential to the overall problem.
We formalize this below.

\subsection{Critical Atoms as Questions Under Discussion (QUD)}
\label{subsec:quds}

\begin{figure}[h!]
\centering
\includegraphics[scale=0.7]{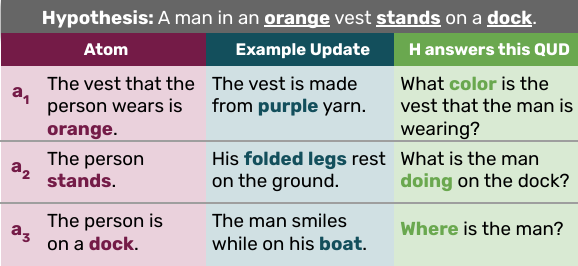}
\caption{Updates ($U$) may act on the same hypothesis $H$ in different ways by targeting different atoms. Here, each $U$ strongly targets a different atom, while having no effect on the other atoms derived from $H$ (e.g. the $U$ in the first row has no effect on $a_3$ in the last row). We refer to the atom(s) which an update most strongly affects as the ``critical'' atom of the $(P, H, U)$ \dnli~example. Critical atoms help identify the \textit{question under discussion} of the example.}
\label{fig:qud-examples}
\end{figure}

As established, updates vary in which atom they most strongly affect.
Consider $H$ and the three updates in Figure~\ref{fig:qud-examples}.
Each $U$ targets a distinct (or \textit{critical}) atom without having an effect on the others.
We formalize this notion by recognizing that hypotheses (or more broadly, sentences in discourse) serve as an answer to a large space of possible questions: all three questions in the right-most column could be answered with $H$.
However, when updates target particular atoms, the \textit{strategy} by which they do so favors a particular question $Q$, making it more likely that $H$ is the answer to $Q$ as opposed to any other question.

\noindent These questions function as \textit{questions under discussion} (QUD), a well-studied linguistic phenomenon~\cite{benz2017questions}. 
The update \textit{``The man smiles on his boat''} has no effect on the atom about vest color and hence does not pick out the QUD associated with that atom.
We call the atoms above the \textit{critical atom} for each corresponding update, as they uniquely pick out particular QUDs over others.
\textbf{Critical atoms correspond to the particular inference or piece of knowledge a defeasible NLI example aims to test.}
As such, we use them as a framework to measure consistency as well as the diversity of the \dsnli~dataset.
This formulation of QUDs relates to question answering-based semantics~\cite{he-etal-2015-question, pyatkin-etal-2021-asking, klein-etal-2022-qasem} in which QA pairs capture semantic information, such as semantic roles.

\paragraph{Identifying Critical Atoms of Updates.} In order to identify the critical atom for a defeasible NLI example, we identify the subset of its \textit{valid} atoms with the strongest labels that match its overall polarity.
The majority of \dsnlitest~examples have one critical atom (Figure~\ref{fig:num-atoms}), but it is possible for example to have multiple if the effect is equally strong, as in row 2 of Table~\ref{table:atom-examples}.

\paragraph{Performance on Critical Atomic Sub-Problems versus Full Examples.}
Across the board, models are stronger on the subset of critical atomic sub-problems than on all atomic problems (Table~\ref{tab:benchmarking-dnli}). 
We hypothesize that LLMs may be better at modeling stronger, \textbf{direct} inferences such as those in critical sub-problems, but may struggle when the effects are \textit{indirect} or weaker.\footnote{For example, $U$ in Row 2 of Table~\ref{table:atom-examples} has no effect on $a_3$, but that effect is slightly strengthened upon mention of a spill in $a_4$, since donning a city uniform is likelier at spill sites.}
Such nuanced distinctions require a robust understanding of the multiple factors that control the underlying inference, a skill that even larger models seem to struggle with.

We also measure the probability that a model correctly predicts the label for the full example \textit{given} that it has correctly (Column 4) and incorrectly (Column 5) solved all critical atomic sub-problems (Table~\ref{tab:benchmarking-dnli}).
Correctly solving all atomic sub-problems is a strong indicator that a model is likely to predict the full problem correctly.
However, some models still have as high as a 75\% probability of predicting the full answer even having incorrectly predicted critical sub-problems, calling into question the robustness of their reasoning process in the face of such inconsistency.

%% file: tables/benchmarking_dnli.tex
\begin{table*}[h!]
\centering
\resizebox{1.5\columnwidth}{!}{
\setlength\tabcolsep{2pt}
\begin{tabular}{lccccc} 
\toprule
 & \begin{tabular}[c]{@{}c@{}}\textbf{Full Example}\\\textbf{Accuracy}\end{tabular} & \begin{tabular}[c]{@{}c@{}}\textbf{Atom}\\\textbf{Accuracy}\end{tabular} & \begin{tabular}[c]{@{}c@{}}\textbf{Critical Atom}\\\textbf{Accuracy}\end{tabular} & \textbf{P(Full \cmark \textbar{} Critical \cmark)} & \textbf{\textbf{P(Full \cmark \textbar{} Critical \xmark)}} \\ 
\midrule
\texttt{llama-3-8b-it} & $80.1$ & $65.3$ & $77.0$ & $90.9$ & $45.9$ \\
\texttt{gpt-3.5-turbo} & $81.5$ & $66.1$ & $76.4$ & $91.8$ & $49.7$ \\
\texttt{gemma-2-9b-it} & $82.3$ & $68.5$ & $72.1$ & $91.9$ & $60.5$ \\
Human~ & $83.6$ & – & – & – & – \\
\texttt{gpt-4o-mini-2024-07-18} & $86.9$ & $74.8$ & $79.4$ & $93.1$ & $66.4$ \\
\texttt{roberta-large} & $87.4$ & $83.4^*$ & $87.8^*$ & $94.0$ & $44.9$ \\
\texttt{llama-3-70b-it} & $88.0$ & $73.8$ & $81.9$ & $93.6$ & $64.5$ \\
\texttt{deberta-v3-large} & $91.1$ & $\mathbf{87.4}^*$ & $\mathbf{91.5}^*$ & $95.0$ & $55.5$ \\
\texttt{gpt-4o-08-06} & $\mathbf{92.6}$ & $77.2$ & $83.5$ & $96.5$ & $75.5$ \\
\bottomrule
\end{tabular}}
\caption{LLM accuracies on full $\delta$-SNLI examples (col. 1), all atomic sub-problems (col. 2), and just critical atom sub-problems (col. 3), using our manual atomic labels. We also report full $\delta$-SNLI example accuracy conditioned on accurate (col. 4) and inaccurate (col. 5) prediction of the corresponding critical atom sub-problem label.}
\label{tab:benchmarking-dnli}
\end{table*}

%% file: sections/55-diversity.tex
\begin{figure}[t!]
\centering
\includegraphics[scale=0.6]{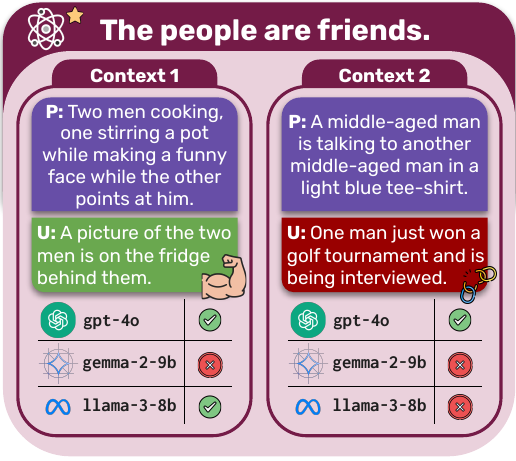}
\caption{Grouping examples by their critical atom(s) allows us to understand under which \textit{contexts} ($P + U$) a model has understood a piece of knowledge. Here, we show two \dnli~examples that evaluate the same atom (top): one that \textit{strengthens} it (left), and one that \textit{weakens} it (right). 
A model that truly understands a fact and the factors that influence it (or, conversely does not) should yield consistently correct or incorrect predictions. However, some models have mixed accuracy among examples targeting the same atom, indicating that they only understand the inference under \textit{some} contexts.}
\label{fig:critical-atom}
\end{figure}

As established, identifying the critical atom(s) of a \dnli~example allows us to pinpoint the knowledge or fact that the example is designed to evaluate (\S\ref{subsec:quds}).
The two different examples in Figure~\ref{fig:critical-atom} share the same critical atom, representing two different contexts under which the model must directly evaluate the fact \textit{``The people are friends.''}
A model correctly predicting whether or not people are friends under some contexts but not others may indicate that it has not fully understood the factors that influence the inference.
Because they contain independent examples, few datasets accommodate measuring a model's \textit{inferential consistency} ($I_C$), or the likelihood that its prediction for a particular inference will remain consistently correct or incorrect under different contexts (here, contexts refers to different $(P, U)$ pairs). 
Correctly drawing an inference under a single context does not guarantee that the model will make a correct prediction for the same inference under a different context.

To quantify this, we group examples in \dsnlitest~by their critical atom and report the inferential consistency of different models.

\paragraph{How many unique critical atoms does \dsnlitest~evaluate?} While \dsnli~contains around 2K test examples and 88K training examples, how many distinct critical atoms underlie those examples? 
We quantify this by identifying semantically equivalent critical atoms in \dsnlitest~by embedding each atom with \texttt{NV-Embed-7B}~\cite{lee2024nv} and then computing pairwise cosine similarity between all critical atoms.
We construct a graph $G$ where nodes correspond to critical atoms and edges are drawn between nodes if their similarity is above a threshold ($\theta=0.75$) and they are bidirectionally entailed as determined by the NLI model in \S\ref{subsec:validate-atoms}.
Finding semantically equivalent groupings then reduces to finding all maximal cliques~\cite{tomita2006worst} in $G$.

The 1,761 \dsnlitest~examples that contain at least one valid atom (from 193 unique $P$-$H$ pairs) contain 429 unique atoms.\footnote{Note that the number of unique atoms is upper-bounded by the \snli~hypotheses that \dsnli~builds upon.}
Following the procedure above yields 349 unique cliques, or \textbf{349 unique \textit{critical} atoms}.\footnote{We also experiment with generating the QUDs (Appendix~\ref{appendix:qud-diversity}) for each critical atom and find that \dsnlitest~evaluates 223 unique QUDs, around an 81\% coverage of the available atoms.}
We bucket examples by these cliques, resulting in groups of examples that share a critical atom (which we call \textit{critical atom buckets}, or just \textit{buckets}, represented as \bucket).
We find that certain critical atoms arise frequently: top critical atoms include \textit{``The others are friends of the person''} (21 examples) or \textit{``The person is a man''} (20 examples).
\paragraph{Measuring Inferential Consistency ($I_C$).} \citet{srikanth2024often} introduce \textit{paraphrastic consistency}, a metric capturing the probability that a model's prediction for two paraphrases of the same NLI problem remain consistent (both incorrect or both correct).
We adapt this metric to compute a model's \textit{inferential consistency}, or the probability that its predictions for two defeasible NLI examples $e_i$ and $e_j$ \textbf{that share the same critical atom} are either both correct or incorrect.
As in \citet{srikanth2024often}, we define two terms: 
\begin{packed_itemize}
  \item $R_{\bucketsmall}$: a discrete random binary variable representing whether a model's prediction for a full \dnli~example is correct (1) or incorrect (0). 
  \item $\theta_{\bucketsmall}$: $\mathds{E}[R_{\bucketsmall}]$, or the average correctness (i.e., accuracy) of the \dnli~examples in a particular critical atom bucket.
\end{packed_itemize}

\noindent For a binary classification task ($y \in \{0, 1\}$) and a model $M$, we define $I_C$ as
\vspace{-0.5em}
\begin{multline}
    I_C = \underbrace{P(M(e_i)= y, M(e_j) = y)}_{\text{prob. of both predictions correct}} + \\
    \underbrace{P(M(e_i) \neq y, M(e_j) \neq y)}_{\text{prob. of both predictions incorrect}}
\end{multline}

\noindent We estimate $I_C$ directly from the accuracies of critical atom buckets as: 
\vspace{-0.5em}
\begin{equation}
    I_C = \mathds{E}[\theta_{\bucketsmall}^2] + \mathds{E}[(1-\theta_{\bucketsmall})^2]
\label{eq:pstay}
\end{equation}

\noindent Note that \dnli~examples can have multiple critical atoms (Figure~\ref{fig:num-atoms}).
In these cases, we divide the weight of the example $e$ across all critical example buckets that share $e$ when computing $\theta_{\bucketsmall}$.
\input{tables/inferential_consistency}

\paragraph{Results.} All models exhibit room for improvement in inferential consistency (Table~\ref{tab:inferential-consistency}), giving us a sense of how well the LLMs we analyze have internalized the critical atomic facts evaluated in \dsnlitest.
One source of inconsistency we observe arises when certain contexts (premise-update pairs) demand more implicit reasoning or background knowledge.
For example, consider two different contexts under which the model must evaluate the same critical atom \textit{``The people are tall''}:

\begingroup
\addtolength\leftmargini{-0.2in}
\begin{quote}
\small

\noindent \colorbox{lightpurple}{\textcolor{darkpurple}{\textbf{Context 1:}}} $P$: Four people standing on a hiking trail in a forest with big tree logs on the ground, $U$: Their long legs step across several logs at once. \\ 
            \colorbox{lightteal}{\textcolor{teal}{\textbf{Context 2}}} $P$: Two men in orange uniforms stand before a train and do some work, $U$: They can easily touch the top of the train.
\end{quote}
\endgroup

\noindent Several models struggle to draw the strengthening inference under the first context, but all models that we analyze successfully draw the strengthening inference under the second.

These analyses help us understand whether models understand pieces of knowledge and the factors that influence them, raising interesting questions about how best to collect updates to increase the coverage of a fact or situation.
Defeasible reasoning systems must be able to deftly modulate inferences in a manner that is sensitive to diverse contexts. 
Future work may leverage the identification of critical atomic sub-problems to nudge annotators toward underrepresented critical atoms or contexts.

%% file: tables/inferential_consistency.tex
\begin{table}[h!]
\centering
\small
\setlength\tabcolsep{2pt}
\begin{tabular}{l c} 
\toprule
\textbf{Model} & \begin{tabular}[c]{@{}c@{}}\textbf{Inferential}\\\textbf{Consistency}\end{tabular} \\ 
\midrule
\texttt{llama-3-8b-it} & $74.5$ \\
\texttt{gpt-3.5-turbo} & $75.8$ \\
\texttt{gemma-2-9b-it} & $76.5$ \\
\texttt{gpt-4o-mini-2024-07-18} & $81.7$ \\
\texttt{roberta-large} & $82.6$ \\
\texttt{llama-3-70b-it} & $83.5$ \\
\texttt{deberta-v3-large} & $87.0$ \\
\texttt{gpt-4o-08-06} & $\mathbf{88.7}$ \\
\bottomrule
\end{tabular}
\caption{Inferential consistency of models ($I_C$) on \dnli~examples. We group examples that share the same critical atom and compute the probability that two examples in the same group were both incorrectly or both correctly predicted by a model.}
\label{tab:inferential-consistency}
\end{table}

%% file: sections/70-conclusion.tex
Hypothesis decomposition allows us to measure the logical consistency of models as well as better understand the structure of non-monotonic reasoning.
We find that labels of atomic sub-problems in defeasible reasoning share a more complex relation to the full problem than in traditional NLI, even within the same label class.
We introduce \textit{critical atoms} as the primary fact evaluated by a defeasible NLI example, enabling us to group examples by shared critical atoms and measure the \textit{inferential consistency} of LLMs across different contexts.

%% file: sections/appendix.tex
\input{tables/annotated_atom_stats}
\input{tables/proportion-completeness}

\section{Atom Generation}
\label{appendix:atom-generation}
\vspace{-3em}
\begin{prompt}[title={Prompt \thetcbcounter: Atom Generation}, label=prompt:atom-generation]
\texttt{\colorbox{lightblue}{Prompt:} You are an expert linguist. You are given a \textbf{sentence}. Generate a list of atomic facts that are strictly logically entailed from the given \textbf{sentence}. Keep each fact independent and self-contained. Each fact should make sense when read on its own. Only write facts that are directly described or supported by the \textbf{sentence}. End your response with [END].\\\\SENTENCE: \{\textbf{sentence}\}\\\\FACTS:}
\end{prompt}

\subsection{Coverage of Generated Atoms}
\label{appendix:coverage-atoms}
Generated atoms must cover the information presented in the hypothesis. 
For example, for the hypothesis $H$ in Figure~\ref{fig:teaser-example}, an atom generation model should produce atoms covering all three pieces of information: at least one mentioning \textit{professional}, \textit{actors}, and \textit{summer}.
Pruning should not reduce the coverage of the atom set.
Here, we estimate the completeness, or coverage, of the generated atoms with respect to the hypothesis of NLI examples. 

\paragraph{Completeness in the case of SNLI.} We note that completeness of atoms only matters in particular cases. 
When the overall $P$-$H$ pair is predicted by the model to be entailed, the logical consistency check does not rely on completeness, as no single atom, if predicted by the model to be entailed, should be predicted as anything other than entailment.
However, there are two possible scenarios of inconsistency when $P-H$ is predicted to be neutral: (a) All atoms were predicted to be entailed by P (this does necessitate ensuring completeness), or (b) one or more atoms is predicted to be a contradiction (this does not require ensuring completeness).
Consistency when $P$-$H$ is predicted to be a contradiction does require checking that all atoms cover the hypothesis.
Table~\ref{tab:proportion-completeness} shows the proportion of examples in \snlitest~that are dependent upon a completeness assumption as per the description for each label above.
These help contexualize our results in Table~\ref{tab:benchmarking-snli} by estimating an upper bound of logical consistency.

To measure completeness in SNLI, we randomly sample 50 SNLI examples and annotate for completeness, and find that in 49/50 examples, all pieces of information from the original hypothesis project into at least one atom. The only example in our random sample that was missing an atom was for the hypothesis shown in Table~\ref{tab:completeness-snli}.

We also sample another set of 50 random examples from \snlitest~where one of our models (\texttt{gpt-4o-2024-08-06}) predicted either contradiction or neutral \textit{and} where consistency failed in order to understand of what percentage of errors are due to actual failures in consistency or are simply due to lack of completeness.
Here, we study the atoms that the model deemed entailed from the hypothesis for each example, and annotate whether those cover all pieces of information in the hypothesis.
In 6/50 examples, \texttt{gpt-4o-2024-08-06} incorrectly judged that at least one of the generated atoms was not entailed by the hypothesis, hence omitting it and causing a completeness issue for the set of atoms over which we measure its logical consistency. 
However, we find that of these six cases, only two examples were missing an atom in the set of generated atoms, again indicating that our \textit{atom generation} process reliably projects all information from the hypothesis into the generated set of atoms.

\paragraph{Completeness in the case of \dnli.} We manually validated all generated 4,079 atomic subproblems in the entire test set of \dsnlitest. 
Lack of completeness is most likely when none of the atoms have a gold label in the direction of the gold label of the overall example, indicating that an atom may be missing.
This happens in only 3\% of examples (70/1761) in \dsnlitest.
We annotate these 70 examples to understand whether or not an atom was indeed missing after our automatic pruning step. 
We find that in only 28 of the 70 examples (representing only ~1\% of \textit{all} examples), at least one atom was missing.
Most of the cases where none of the atoms have a gold label in the direction of the gold label of the overall example are cases where the original example is flawed in some way (hinges on some stereotype, ambiguous language, faulty reasoning, or the original crowdworker who wrote the update did not understand the instructions) and our annotations do not propagate the flawed assumptions or reasoning from the original example. Table~\ref{tab:completeness} shows an example.
The original gold label propagates a stereotype or attitude towards janitors (i.e it is less likely they have an important meeting if they are a janitor). We choose not to propagate such attitudes or stereotypes in our annotation, hence, all atoms have the ``no effect'' label.

\begin{table}[t]
    \centering
    \resizebox{\columnwidth}{!}{%
    \begin{tabular}{ll}
        \toprule
        \rotatebox[origin=c]{90}{\textbf{Example}} &
        \makecell*[{{p{10.7cm}}}]{
            \colorbox{lightpurple}{\textcolor{darkpurple}{\textbf{Premise:}}} A female within the foreground is heading towards a large white colored pillar that is apart of a large building with people are loitering or waiting on the steps of said building. \\ 
            \colorbox{lightteal}{\textcolor{teal}{\textbf{Hypothesis}}} The woman has an important meeting today in the building. \\
            \colorbox{lightorange}{\textcolor{burntorange}{\textbf{Update}}} The woman is wearing a janitor's uniform (\textit{weakener})
        }\\
        
        \midrule
        \rotatebox[origin=c]{90}{\textbf{\textcolor{coolgreen}{Atoms}}}
        &\makecell*[{{p{10cm}}}]{
        $a_1$: The person has something (0, no effect).\\
        $a_2$: The thing the person has is a meeting. (0, no effect) \\
        $a_3$: The meeting is important (0, no effect) \\
        $a_4$: The meeting is today (0, no effect) \\
        $a_5$: The meeting is in a building (0, no effect) \\
}\\

        \bottomrule
    \end{tabular}}
    \caption{An example of a lack of completeness in generated atoms.}
    \label{tab:completeness}
\end{table}

\begin{table}[t]
    \centering
    \resizebox{\columnwidth}{!}{%
    \begin{tabular}{ll}
        \toprule
        \rotatebox[origin=c]{90}{\textbf{Hypothesis}} &
        \makecell*[{{p{10.7cm}}}]{
            The man and woman are going to a movie in the city.
        }\\
        
        \midrule
        \rotatebox[origin=c]{90}{\textbf{\textcolor{coolgreen}{Atoms}}}
        &\makecell*[{{p{10cm}}}]{
        $a_1$: There are two people. \\
        $a_2$: There are two people who are a man and a woman. \\
        $a_3$: There are people going. \\
        $a_4$: There are people going to a movie. \\
        $a_5$: There is a destination for the people going. \\
        $a_6$: The destination is a movie. \\
        $a_7$: There is a city. \\
        $a_8$: The people are going to the city.
}\\

        \bottomrule
    \end{tabular}}
    \caption{An example of a lack of completeness in generated atoms for SNLI. Here the missing atom is \textit{``The movie is in the city.''}}
    \label{tab:completeness-snli}
\end{table}

\section{SNLI}
\begin{prompt}[title={Prompt \thetcbcounter: Traditional NLI}, label=prompt:nli]
\texttt{\colorbox{lightblue}{Prompt:} You will be given a \textbf{premise} and a \textbf{hypothesis} about that \textbf{premise}. You need to decide whether the \textbf{hypothesis} is entailed by the \textbf{premise} by choosing one of the following answers: 'e': The \textbf{hypothesis} follows logically from the information contained in the \textbf{premise}. 'c': The \textbf{hypothesis} is logically false from the information contained in the \textbf{premise}. 'n': It is not possible to determine whether the \textbf{hypothesis} is true or false without further information. Read the \textbf{premise} and \textbf{hypothesis} and select the correct answer from the three answer labels (e, n, c). Also provide a single line of explanation in a new line. End your response with [END] and output nothing after.\\\\Premise: \{premise\}\\Hypothesis:\{hypothesis\}\\\\Is the hypothesis entailed by, contradicted by, or neutral with respect to the premise?}
\end{prompt}

\subsection{Inconsistency in SNLI}
\label{appendix:inconsistency-analysis}
We analyze the set of randomly selected sample of 50 examples from Appendix~\ref{appendix:coverage-atoms} where one of our models (\texttt{gpt-4o-2024-08-06}) was logically inconsistent.
We find that inconsistencies arise for a number of reasons, some of which we discuss here.

\paragraph{Misjudgment of atom entailment from the hypothesis.} Measuring logical consistency in SNLI examples happens in two phases (1) the model determines whether the atom entails from the hypothesis, and (2) the model predicts the relationship between the premise and an atom. Depending on its strength, we find that there are inconsistencies that arise from the model’s misjudgment of atom entailment.
For example: 

\begingroup
\addtolength\leftmargini{-0.2in}
\begin{quote}
\small

\noindent \colorbox{lightpurple}{\textcolor{darkpurple}{\textbf{Premise:}}} \textit{A man on a bicycle, wearing cycle gear, riding at a fast past down paved trail surrounded by tree's and grass.} \\ 
            \colorbox{lightteal}{\textcolor{teal}{\textbf{Hypothesis}}} \textit{The man is riding on the sidewalk.}\\
            \colorbox{lightorange}{\textcolor{burntorange}{\textbf{Generated Atoms}}} \textit{($a_1$) There is a person. ($a_2$) There is a person who is a man.
 ($a_3$) There is a person riding. ($a_4$) There is a person riding on something. ($a_5$) The thing the person is riding on is a sidewalk.}
\end{quote}
\endgroup

 \noindent Based on the full premise and hypothesis, \texttt{gpt-4o-2024-08-06} predicted that the hypothesis was \textit{neutral} in relation to the premise. 
 However, in its atomic predictions between the premise and each atom, the model incorrectly determined that the last atom was not entailed by the hypothesis, and hence it was not included in the set of atoms that were used to measure consistency. It judged all other atoms as entailed from the premise, hence leading to the model behaving inconsistently in this example.

\paragraph{Use of hypernyms in atoms.} We make sure to include hypernyms of entities in the set of atomic facts. 
For example, \textit{``the man dances''} is decomposed into \textit{``there is a person'', ``the person is a man'', ``the person dances''}. 
In some cases, models have difficulty on premise-atom judgments where the premise uses the hyponym (``man'') and the hypothesis uses the hypernym (``person'').

\paragraph{Out of domain syntactic constructions.} The syntactic structures of many of our atoms differ, in some cases significantly, from the constructions in the original dataset. For example \textit{``the thing the person is eating is a sandwich''} is a pseudo-cleft construction that is very rare in the SNLI dataset. 
As such, some sentences are out of domain for encoder models that were trained on SNLI or prompt-based models that have inadvertently seen SNLI training data in pretraining corpora.

\paragraph{Weaker effects based on annotation elicitation.} Both SNLI and defeasible NLI are datasets created by crowdworkers writing hypotheses and updates conditioned on a label. 
One of the consequences of this process is that hypotheses and updates tend to strongly express the desired label.
In contrast, atoms tend to express labels in a softer way, and their effects often compound when taken together.
As such, these atoms may be out of distribution as compared to hypotheses that express the label with a higher magnitude.
Since the effect of each atom is lighter than when they are taken together in a full sentence, atomic judgments are sometimes much more difficult than overall judgments.

\paragraph{Co-reference Effects.} In some cases, atomic generations remove some of the implicit co-reference between the hypothesis and premise. 
We observe that changing the co-reference in atoms can result in inconsistencies between the overall example and each premise-atom judgment.

\section{Defeasible NLI}

\begin{prompt}[title={Prompt \thetcbcounter: Defeasible Inference}, label=prompt:defeasible]
\texttt{\colorbox{lightblue}{Prompt:} You are a reasoning system. You are given a description of a \textbf{situation} and a \textbf{hypothesis} about that \textbf{situation} that may or may not be true. Given some more \textbf{evidence} about the \textbf{situation}, output 'more' if the \textbf{hypothesis} seems more likely to be true after learning the \textbf{evidence}, or output 'less' if the \textbf{hypothesis} seems less likely to be true after learning the \textbf{evidence}. Also provide a single line of explanation in a new line. End your response with [END] and output nothing after.\\ \\Situation: \{context\}\\\\Hypothesis: \{hypothesis\}\\Evidence: \{evidence\}\\\\Does the evidence make the hypothesis about the situation more or less likely to be true?}
\end{prompt}

\begin{prompt}[title={Prompt \thetcbcounter: Defeasible Inference Atoms}, label=prompt:defeasible-atom]
\texttt{\colorbox{lightblue}{Prompt:} You are a reasoning system. You are given a description of a \textbf{situation} and a \textbf{hypothesis} about that \textbf{situation} that may or may not be true. Given some more \textbf{evidence} about the \textbf{situation}, output 'more' if the \textbf{hypothesis} seems more likely to be true after learning the \textbf{evidence}, output 'less' if the \textbf{hypothesis} seems less likely to be true after learning the \textbf{evidence}, or output 'none' if the likelihood of the \textbf{hypothesis} remains unchanged after learning the \textbf{evidence}. Also provide a single line of explanation in a new line. End your response with [END] and output nothing after.\\ \\Situation: \{context\}\\\\Hypothesis: \{hypothesis\}\\Evidence: \{evidence\}\\\\Does the evidence make the hypothesis about the situation more or less likely to be true?}
\end{prompt}

\begin{figure}[t!]
\centering
\includegraphics[scale=0.5]{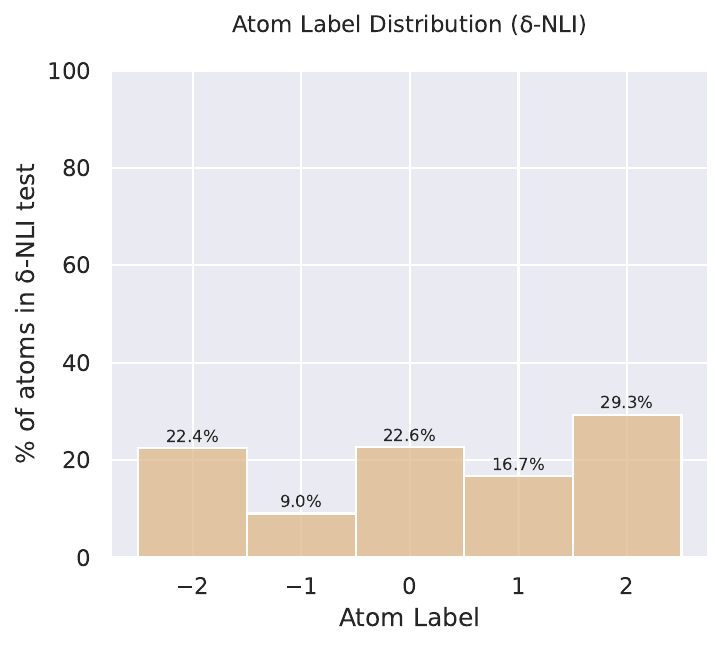}
\caption{Distribution of fine-grained labels across all atoms in \dsnlitest.}
\label{fig:atom-label-hist}
\end{figure}

\begin{figure}
\centering
\includegraphics[scale=0.6]{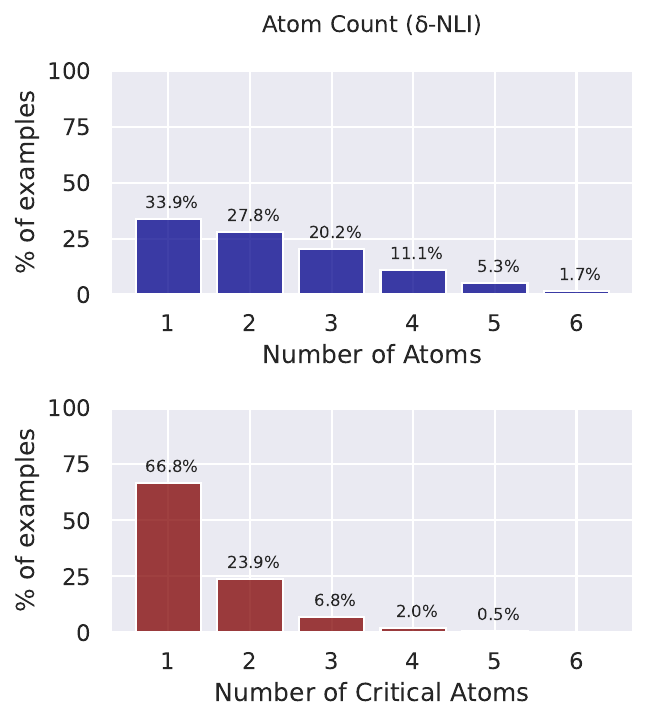}
\caption{Number of atoms (top) and number of critical atoms (bottom) per example in \dsnlitest.}
\label{fig:num-atoms}
\end{figure}

\begin{figure}
\centering
\includegraphics[scale=0.5]{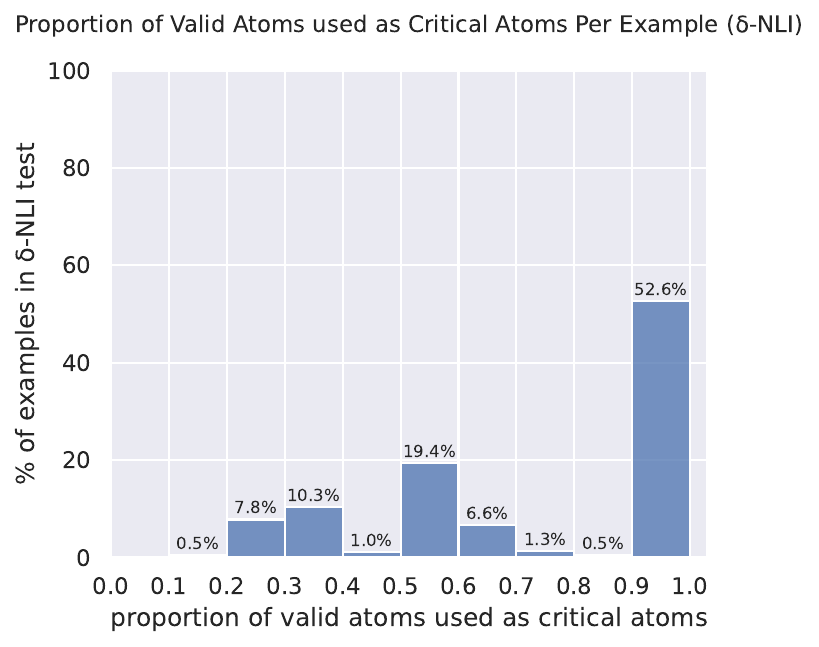}
\caption{Proportion of valid atoms used as critical atoms per update in \dsnlitest. 52\% of updates target all valid possible atoms at once in a single update sentence.}

\label{fig:critical-atom-usage}
\end{figure}

\section{QUD Generation for Understanding Diversity}
\label{appendix:qud-diversity}

\input{tables/qud_generation}

In order to generate QUDs from critical atom propositions, we use \texttt{gpt-4o-2024-08-06} with a temperature of 0.01 and 15 exemplars of critical atom to QUD mappings along with Prompt~\ref{prompt:qud-gen}.
For example, the critical atom \textit{The dog is brown.} translates to the QUD \textit{What color is the dog?} Exemplars include critical atoms that do not use generics (e.g.``girl'') mapped to QUDs that use generics (``person''). This is to make sure that similar inferences are correctly mapped to the same QUD. Table~\ref{tab:quds-generated} shows examples of QUDs generated by the model.
We validate all generated QUDs by asking an external annotator if the QUD can be answered by the critical atom sentence, and find that in 92\% of cases, the QUD generated by \texttt{gpt4o} correctly captures the critical atom.

\begin{prompt}[title={Prompt \thetcbcounter: QUD Generation}, label=prompt:qud-gen]
\texttt{\colorbox{lightblue}{Prompt:} You are an expert linguist. Given a short \textbf{sentence}, generate a \textbf{question} that is answered by the sentence. Read the whole sentence carefully before generating the question.\\\textbf{Sentence}: {critical 
atom}\\\textbf{Question}:}
\end{prompt}

\section{Validation Instructions}
\label{appendix:instructions}
Since atomic sub-problems mirror the original validation task for defeasible inference, we use the instructions provided to annotators from \citet{rudinger-etal-2020-thinking} to ensure alignment.

%% file: tables/annotated_atom_stats.tex
\begin{table}
\centering
\resizebox{\columnwidth}{!}{
\begin{tabular}{lcc} 
\toprule
& \textbf{\snli} & \textbf{\dsnli}  \\ 
\midrule
\textbf{\# test examples~}   & 1000 & 1837  \\
\textbf{label distr}&  36/32/32\% (E/C/N)  &  50/50\% (S/W) \\
\textbf{\# unique hypotheses}  &  1000      & 203   \\
\textbf{\# unique generated atoms}  & 3263  &  475   \\
\textbf{\# unique valid atoms}   & --            & 429 \\
\textbf{\# unique atomic sub-problems}   & 3263            & 4079 \\
\>\>\>\% strengtheners     & --            & 46.0            \\
\>\>\>\%  weakeners  & --            & 31.4            \\
\>\>\>\%  no effect   & --            & 22.6               \\
\textbf{\textbf{\textbf{\textbf{~\# unique critical atoms}}}} & --            & 349   \\
\bottomrule
\end{tabular}}
\caption{Dataset statistics for both \snli~and \dsnli.}
\label{tab:dataset-sizes}
\end{table}

%% file: tables/proportion-completeness.tex
\begin{table}[h!]
\centering
\small
\setlength\tabcolsep{2pt}
\begin{tabular}{l c} 
\toprule
\textbf{Model} & \textbf{Proportion} \\ 
\midrule
\texttt{llama-3-8b-it} & $32.0$ \\
\texttt{gpt-3.5-turbo} & $31.1$ \\
\texttt{gemma-2-9b-it} & $25.5$ \\
\texttt{gpt-4o-mini-2024-07-18} & $34.5$ \\
\texttt{llama-3-70b-it} & $33.3$ \\
\texttt{gpt-4o-08-06} & $27.6$ \\
\bottomrule
\end{tabular}
\caption{Proportion of examples in \snlitest~(1000 examples) depending on a completeness assumption of atom generation.}
\label{tab:proportion-completeness}
\end{table}

%% file: tables/qud_generation.tex
\begin{table*}
\small
\centering
\begin{tabular}{ll} 
\toprule
\multicolumn{1}{c}{\textbf{Critical Atom}} & \multicolumn{1}{c}{\textbf{QUD}} \\ 
\hline
The kids are waiting. & What are the people doing? \\
The dog is brown. & What color is the dog? \\
The thing the grandpa is wearing is a shirt. & What is the person wearing? \\
The two women are sisters. & What is the relationship between the two people? \\
The bike is brand new. & What is the condition of the object? \\
The girls are posing. & What are the people doing? \\
\bottomrule
\end{tabular}
\caption{QUDs generated from critical atoms in \dsnlitest by \texttt{gpt-4o-2024-08-06}.}
\label{tab:quds-generated}
\end{table*}